%% file: main.tex
\documentclass[10pt,twocolumn,letterpaper]{article}

\usepackage{iccv}              


\definecolor{iccvblue}{rgb}{0.21,0.49,0.74}
\usepackage[pagebackref,breaklinks,colorlinks,allcolors=iccvblue]{hyperref}

\def\paperID{13946} 
\def\confName{ICCV}
\def\confYear{2025}

\title{Video Virtual Try-on with Conditional Diffusion Transformer Inpainter}

\author{Cheng Zou\thanks {Equal contributions.}\ , 
Senlin Cheng$^*$, 
Bolei Xu, 
Dandan Zheng,\\
Xiaobo Li,
Jingdong Chen,
Ming Yang\\
Ant Group\\
{\tt\small \{wuyou.zc,senlin.csl,shishi.yl,yuandan.zdd,xiaobo.lixb}\\ 
{\tt\small jingdongchen.cjd,m.yang\}@antgroup.com }
}

\begin{document}

\maketitle

\begin{figure*}[htp]
    \centering

    \begin{subfigure}[b]{0.14\textwidth}
        \centering
        \includegraphics[width=\textwidth]{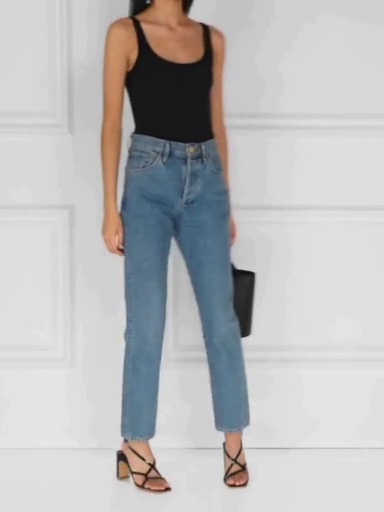}
    \end{subfigure}
    \hspace{0.0\textwidth}
    \begin{subfigure}[b]{0.14\textwidth}
        \centering
        \includegraphics[width=\textwidth]{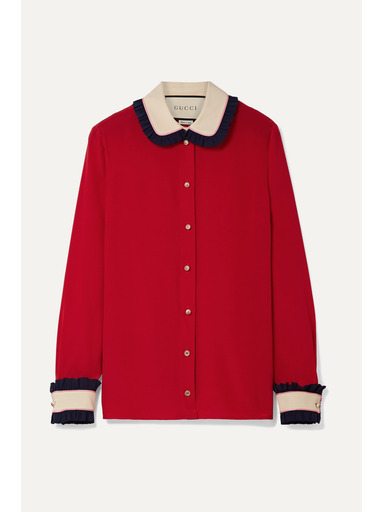}
    \end{subfigure}
    \hspace{0.0\textwidth}
    \begin{subfigure}[b]{0.14\textwidth}
        \centering
        \includegraphics[width=\textwidth]{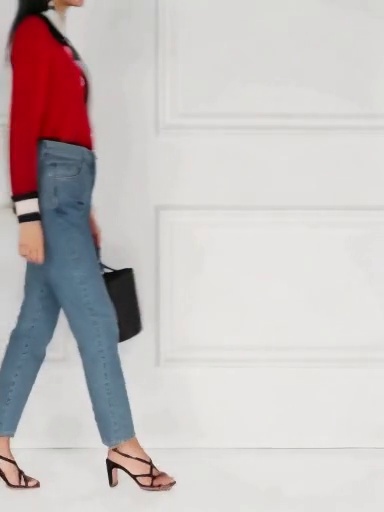} 
    \end{subfigure}
    \hspace{0.0\textwidth}
    \begin{subfigure}[b]{0.14\textwidth}
        \centering
        \includegraphics[width=\textwidth]{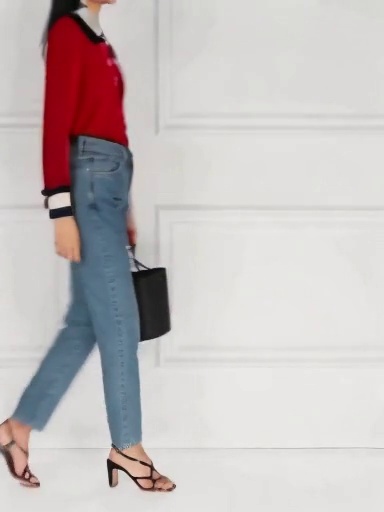} 
    \end{subfigure}
    \hspace{0.0\textwidth}
    \begin{subfigure}[b]{0.14\textwidth}
        \centering
        \includegraphics[width=\textwidth]{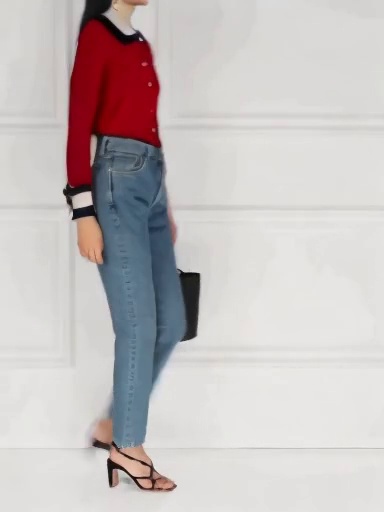} 
    \end{subfigure}
    \hspace{0.0\textwidth}
    \begin{subfigure}[b]{0.14\textwidth}
        \centering
        \includegraphics[width=\textwidth]{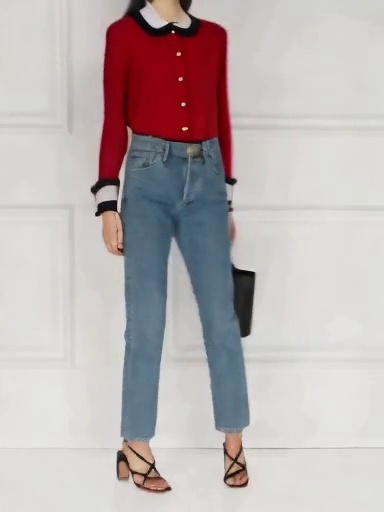} 
    \end{subfigure}
    \begin{subfigure}[b]{0.14\textwidth}
        \centering
        \includegraphics[width=\textwidth]{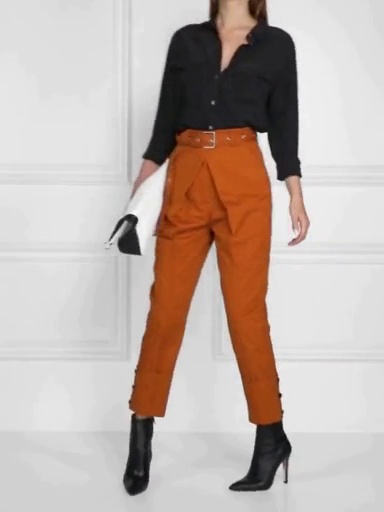}
    \end{subfigure}
    \hspace{0.0\textwidth}
    \begin{subfigure}[b]{0.14\textwidth}
        \centering
        \includegraphics[width=\textwidth]{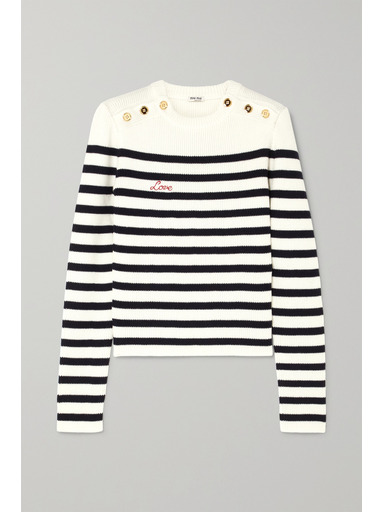}
    \end{subfigure}
    \hspace{0.0\textwidth}
    \begin{subfigure}[b]{0.14\textwidth}
        \centering
        \includegraphics[width=\textwidth]{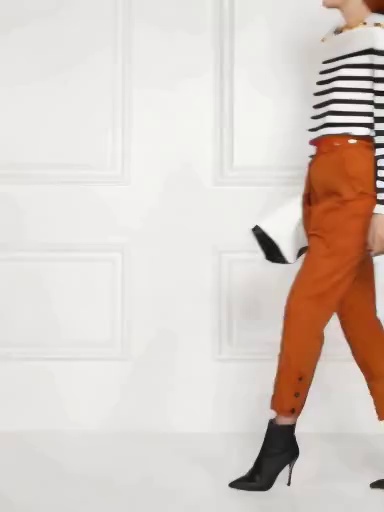} 
    \end{subfigure}
    \hspace{0.0\textwidth}
    \begin{subfigure}[b]{0.14\textwidth}
        \centering
        \includegraphics[width=\textwidth]{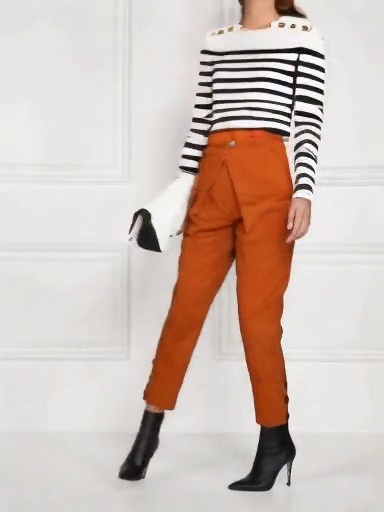} 
    \end{subfigure}
    \hspace{0.0\textwidth}
    \begin{subfigure}[b]{0.14\textwidth}
        \centering
        \includegraphics[width=\textwidth]{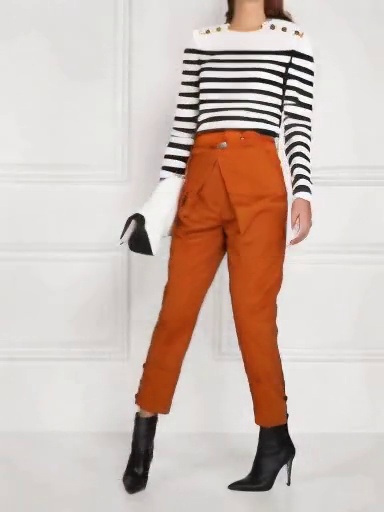} 
    \end{subfigure}
    \hspace{0.0\textwidth}
    \begin{subfigure}[b]{0.14\textwidth}
        \centering
        \includegraphics[width=\textwidth]{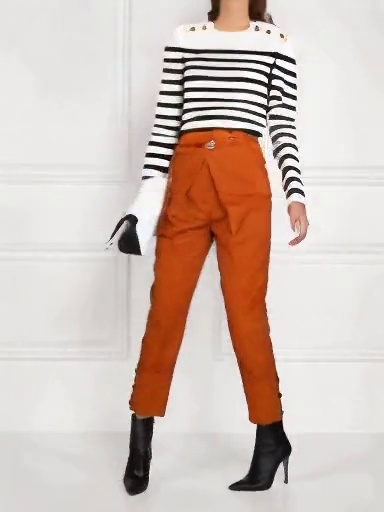} 
    \end{subfigure}

    \begin{subfigure}[b]{0.14\textwidth}
        \centering
        \includegraphics[width=\textwidth]{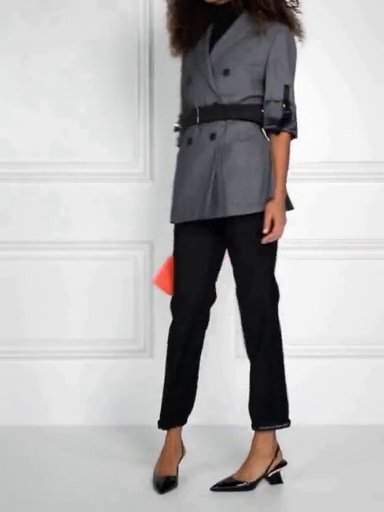}
    \end{subfigure}
    \hspace{0.0\textwidth}
    \begin{subfigure}[b]{0.14\textwidth}
        \centering
        \includegraphics[width=\textwidth]{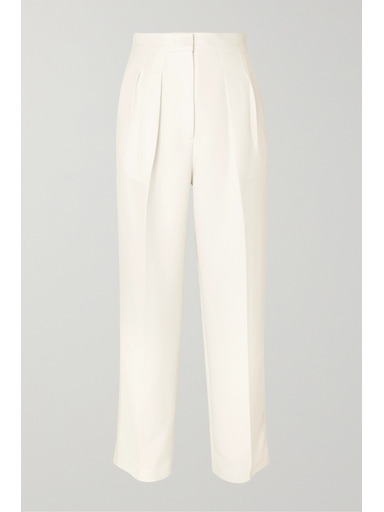}
    \end{subfigure}
    \hspace{0.0\textwidth}
    \begin{subfigure}[b]{0.14\textwidth}
        \centering
        \includegraphics[width=\textwidth]{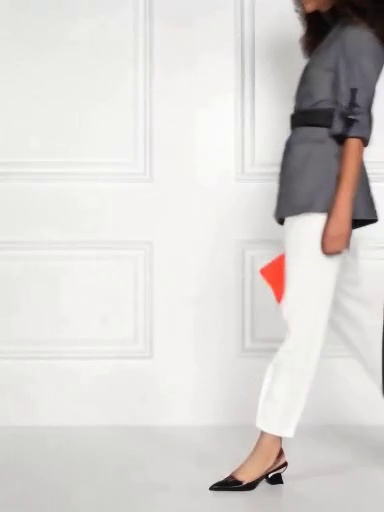} 
    \end{subfigure}
    \hspace{0.0\textwidth}
    \begin{subfigure}[b]{0.14\textwidth}
        \centering
        \includegraphics[width=\textwidth]{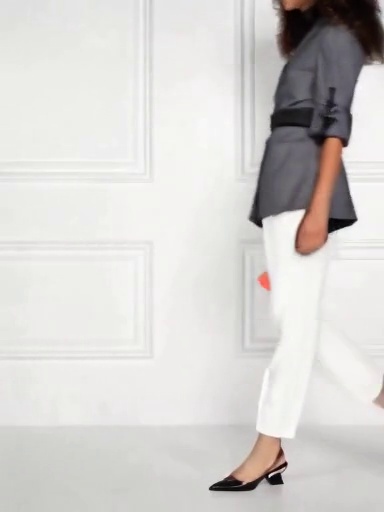} 
    \end{subfigure}
    \hspace{0.0\textwidth}
    \begin{subfigure}[b]{0.14\textwidth}
        \centering
        \includegraphics[width=\textwidth]{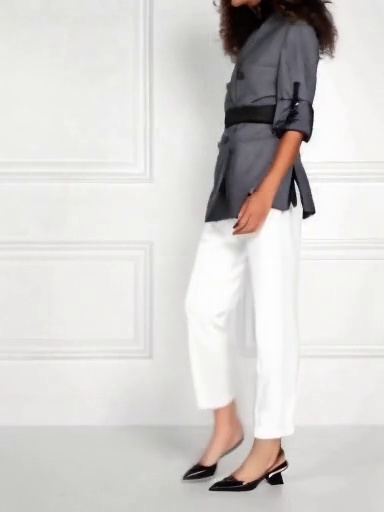} 
    \end{subfigure}
    \hspace{0.0\textwidth}
    \begin{subfigure}[b]{0.14\textwidth}
        \centering
        \includegraphics[width=\textwidth]{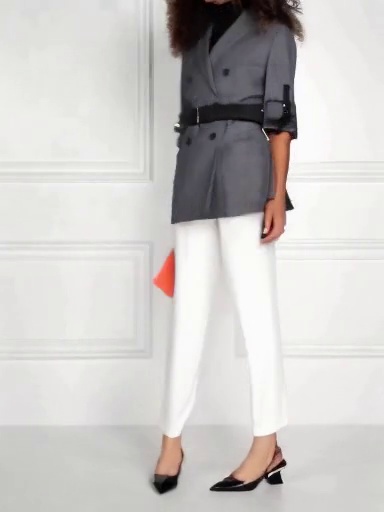} 
    \end{subfigure}
    \begin{subfigure}[b]{0.14\textwidth}
        \centering
        \includegraphics[width=\textwidth]{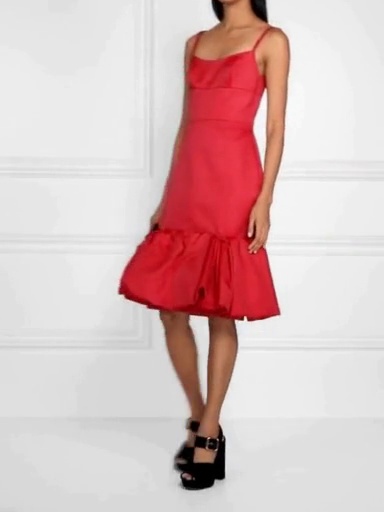}
    \end{subfigure}
    \hspace{0.0\textwidth}
    \begin{subfigure}[b]{0.14\textwidth}
        \centering
        \includegraphics[width=\textwidth]{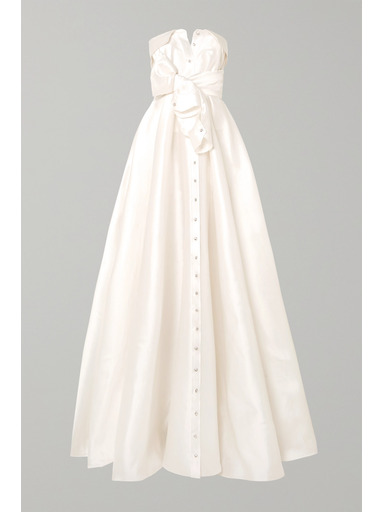}
    \end{subfigure}
    \hspace{0.0\textwidth}
    \begin{subfigure}[b]{0.14\textwidth}
        \centering
        \includegraphics[width=\textwidth]{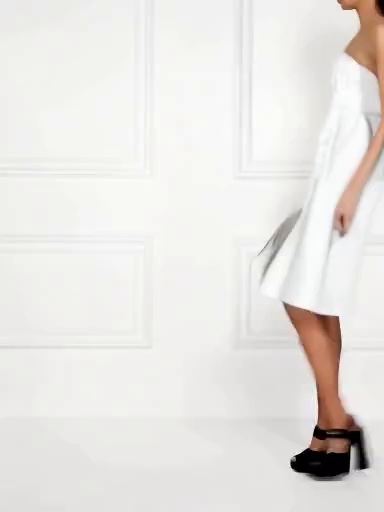} 
    \end{subfigure}
    \hspace{0.0\textwidth}
    \begin{subfigure}[b]{0.14\textwidth}
        \centering
        \includegraphics[width=\textwidth]{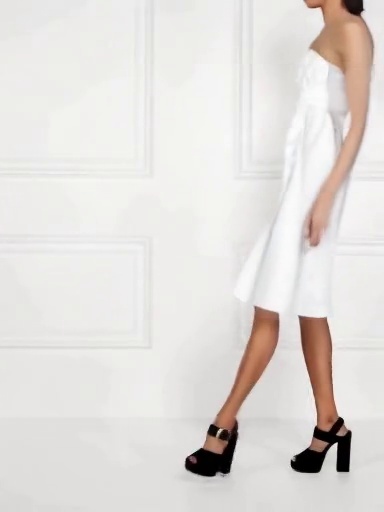} 
    \end{subfigure}
    \hspace{0.0\textwidth}
    \begin{subfigure}[b]{0.14\textwidth}
        \centering
        \includegraphics[width=\textwidth]{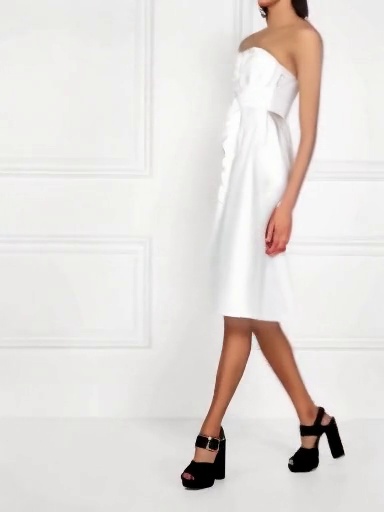} 
    \end{subfigure}
    \hspace{0.0\textwidth}
    \begin{subfigure}[b]{0.14\textwidth}
        \centering
        \includegraphics[width=\textwidth]{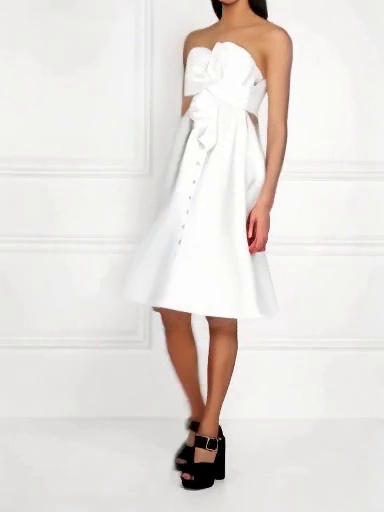} 
    \end{subfigure}

    \caption{Results of ViTI (Video Try-on Inpainter). The first two columns provide the input. ViTI performs well on garment detail preservation and spatial-temporal consistency, and also supports try-on for different clothing types, such as tops, pants, and skirts.}
    \label{fig:first_image}  
\end{figure*}

\input{sec/0_abstract}    
\input{sec/1_intro}

\input{sec/2_formatting}

\input{sec/3_finalcopy}
{
    \small
    \bibliographystyle{ieeenat_fullname}
    \bibliography{main}
}

\end{document}

%% file: sec/0_abstract.tex
\begin{abstract}
Video virtual try-on aims to naturally fit a garment to a target person in consecutive video frames. It is a challenging task, on the one hand, the output video should be in good spatial-temporal consistency, on the other hand, the details of the given garment need to be preserved well in all the frames. Naively using image-based try-on methods frame by frame can get poor results due to severe inconsistency. Recent diffusion-based video try-on methods, though very few, happen to coincide with a similar solution: inserting temporal attention into image-based try-on model to adapt it for video try-on task, which have shown improvements but there still exist inconsistency problems. In this paper, we propose ViTI (Video Try-on Inpainter), formulate and implement video virtual try-on as a conditional video inpainting task, which is different from previous methods. In this way, we start with a video generation problem instead of an image-based try-on problem, which from the beginning has a better spatial-temporal consistency. Specifically, at first we build a video inpainting framework based on Diffusion Transformer with full 3D spatial-temporal attention, and then we progressively adapt it for video garment inpainting, with a collection of masking strategies and multi-stage training. After these steps, the model can inpaint the masked garment area with appropriate garment pixels according to the prompt with good spatial-temporal consistency. Finally, as other try-on methods, garment condition is added to the model to make sure the inpainted garment appearance and details are as expected. Both quantitative and qualitative experimental results show that ViTI is superior to previous works.
\end{abstract}

%% file: sec/1_intro.tex
\section{Introduction}
\label{sec:intro}

Virtual try-on provides a practical commercial scene in online shopping that the customers can try on some preferred clothes without touching or physically trying on them before purchasing, which can reduce the probabilities of refunds, exchanges and returns, and in the end improve the shopping experience and stimulate future shoppings. Also, it provides a low cost trail and error solution for designers, which can improve the efficiency of costume design.

Intuitively, video virtual try-on is much more user-friendly than image virtual try-on, because it can bring more impressive information for the customer, i.e., the figure from different angles, the dynamic movements of the clothes when trying on his/her body. However, previous works mainly focus on image virtual try-on~\cite{kim2023stableviton,zhu2023tryondiffusion,lee2022high,morelli2022dress,ge2021parser,choi2021viton,issenhuth2020not,wang2018toward,han2017viton,gou2023taming,morelli2023ladi,baldrati2023multimodal}, very few studies have attempted video virtual try-on~\cite{fang2024vivid,xu2024tunnel,jiang2022clothformer}, so the current ability is far from realistic applications. 

On the way to powerful video virtual try-on system, there are two main obstacles facing us, one is to precisely preserve the details of the garment, the other is to ensure the spatial-temporal consistency of the generated video. Recent state-of-the-art diffusion-based video virtual try-on methods~\cite{fang2024vivid,xu2024tunnel} happen to coincide with a similar solution, that is, training an image-based try-on model at first to acquire the basic try-on ability, and then insert temporal module to improve temporal consistency. In this first-image-then-video paradigm, the image part is usually a main U-Net pretrained on a large number of image data, while the video part is just temporal attention layer after the spatial attention in the main U-Net. As is known, the temporal attention layer is designed to conduct self-attention on features of the same spatial position across different frames, which is more like a temporal smoother, though efficient but with poor ability for spatial-temporal modeling. So these methods still suffer from inconsistency problems. The main difference between video and image virtual try-on lies in the demand of spatial-temporal consistency, which is also the fundamental of natural videos. The above mentioned first-image-then-video methods, can be interpreted as image-based methods decorated with limited temporal filters or smoothers, naturally have inconsistency problems. 

In this paper, we propose ViTI (Video Try-on Inpainter) and formulate video virtual try-on as a conditional video inpainting task, which is  different from previous methods. In this way, we start with a video generation problem instead of an image-based try-on problem, which from the beginning will have a better spatial-temporal consistency. Specifically, we first build a video inpainting framework based on Diffusion Transformer~\cite{pku_yuan_lab_and_tuzhan_ai_etc_2024_10948109} with full 3D spatial-temporal attention. To train this video inpainting model, we design a collection of masking strategies, such as time-invariant box mask, time-variant box mask, instance-level mask, etc., to make the model robust when given a variety of mask shapes. As far as we know, we are the first to implement video inpainting with full 3D attention Diffusion Transformer. We choose full 3D attention because it has a strong spatial-temporal modeling ability for video generation, let alone video inpainting and video virtual try-on. When the video inpainting framework is ready, we progressively adapt it for video garment inpainting task with multi-stage training. And to train the model, we further collect a human-centric dataset with 51,278 video clips for video try-on pretraining, whose clothes are segmented and masked with pretrained segmentation model~\cite{khirodkar2024sapiens}, which is the first large scale dataset for video try-on pretraining. After these steps, the model can inpaint the masked garment area with appropriate garment pixels according to the prompt with good spatial-temporal consistency. Finally, garment condition is added to make sure the inpainted garment appearance and details are as expected. During training, a temporal consistency loss for diffusion model is proposed to explicitly constrain the difference between consecutive latent frames to get better temporal consistency.

\section{Related Work}
\label{sec:related_work}

\subsection{Image Virtual Try-on}
Though GAN-based image virtual try-on methods~\cite{lee2022high,morelli2022dress,ge2021parser,choi2021viton,issenhuth2020not,wang2018toward,han2017viton} are pioneers, we pay more attention to the recent diffusion-based methods~\cite{gou2023taming,morelli2023ladi,baldrati2023multimodal,kim2023stableviton,zhu2023tryondiffusion}. The key problem in image virtual try-on lies in the garment detail preservation and warping. 
DCI-VTON~\cite{gou2023taming} 
proposes an exemplar-based approach that leverages a warping module to guide the diffusion model, which helps to preserve the local details. 
LaDI-VITON~\cite{morelli2023ladi} treats garment image features as a set of pseudo-word token embeddings to maintain the texture and details of the garment. 
StableVITON~\cite{kim2023stableviton} proposes zero cross-attention blocks, which not only preserve the garment details by learning the semantic correspondence but also generate high-fidelity images by utilizing the inherent knowledge of the pretrained model in the warping process. OOTDiffusion~\cite{xu2024ootdiffusion} proposes outfitting fusion in the self-attention layers of the denoising UNet to precisely align the garment features with the target human body. TryOnDiffusion~\cite{zhu2023tryondiffusion} employs parallel-UNet to preserve garment details and warp the garment for significant pose and body change in a single network.
These methods have made great progress in image try-on to generate high-fidelity images, but when applied to video try-on frame by frame there will be poor results due to severe inter-frame inconsistency.

\subsection{Video Virtual Try-on}

One of the most important things for video virtual try-on is to ensure spatial-temporal consistency. FW-GAN~\cite{dong2019fw} uses a flow-guided fusion module to warp the past frames to generate coherent video, a warping net to warp clothes, and a parsing constraint loss to alleviate the misalignment problem, which leading to good temporal consistency and visual quality. ClothFormer~\cite{jiang2022clothformer} proposes an appearance-flow tracking module that utilizes ridge regression and optical flow correction to smooth the dense flow sequence to generate a temporally consistent warped garment sequence. MV-TON~\cite{zhong2021mv} generates a coarse try-on result at first, and then adopts a memory refinement module to save the previously generated frames for the following frame generation. 
Tunnel Try-on~\cite{xu2024tunnel} leverages the Kalman filter to construct smoother crops in the focus tunnel and inject position embedding of the tunnel  to improve continuity, also it zooms in on the focus tunnel region to better preserve the fine details of the garment. ViViD~\cite{fang2024vivid} proposes a diffusion-based framework and insert temporal modules into the text-to-image stable diffusion model for coherent video synthesis.
The recent state-of-the-art diffusion-based video virtual try-on methods still suffer from inconsistency problems. We humbly hypothesize that this is mainly due to the limitation of the 1D temporal attention module and the absence of modern video generation improvements. In this paper, we formulate video virtual try-on as a conditional video inpainting task and implement it with a modern video generation paradigm, which obviously has a better spatial-temporal consistency.


\begin{figure*}[tp]
	\centering
	\includegraphics[width=0.80\textwidth]{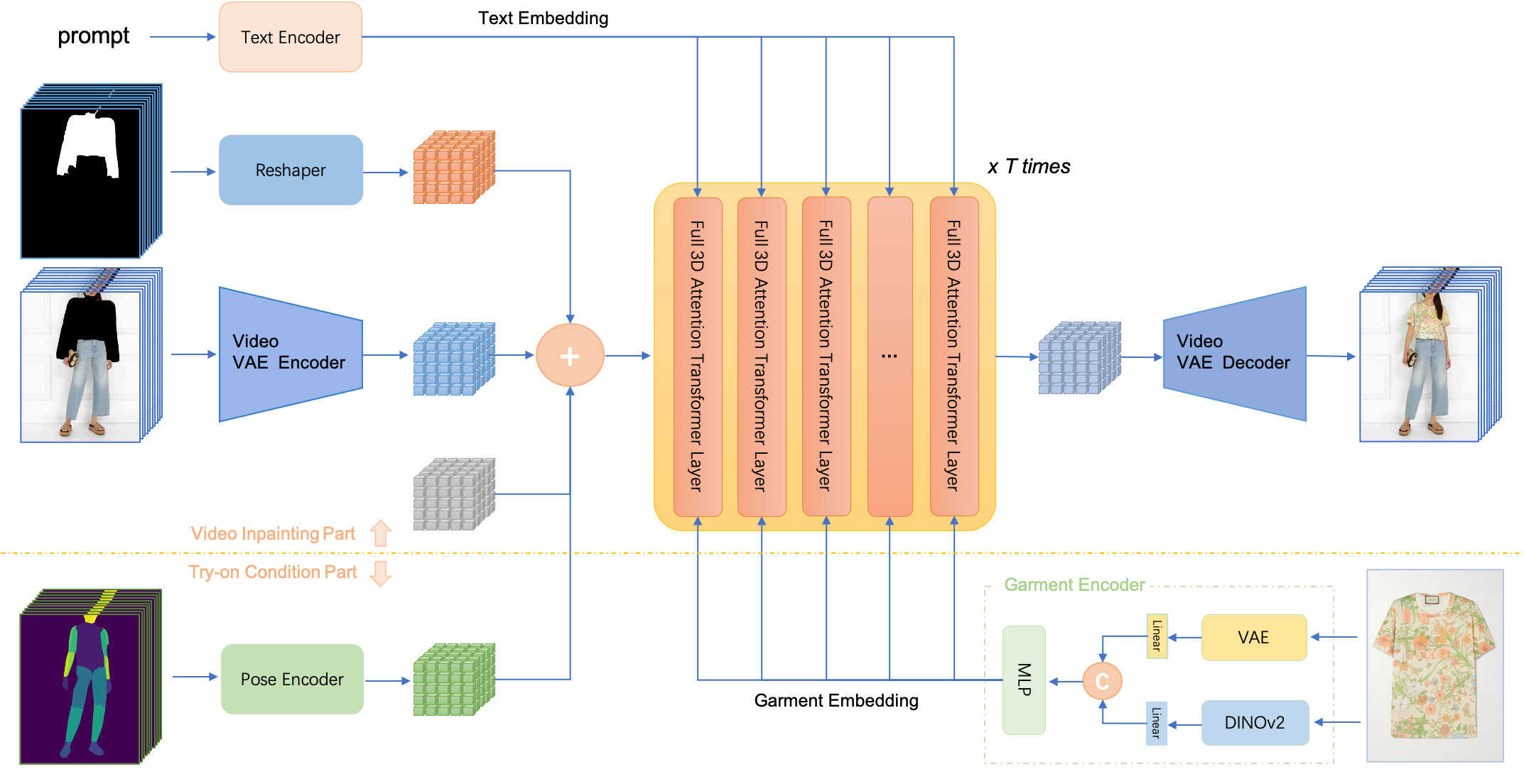}
	\caption{Overall architecture. The main framework of ViTI (Video Try-on Inpainter) consists of two parts, the video inpainting part, and the try-on condition part. For video inpainting, the mask video through reshaper, the agnostic video through video VAE encoder, are added with the diffused noise in latent space to feed into Diffusion Transformer (DiT) as input. After a $T$-step denoising, the output tokens of DiT is decoded to pixel space through a video VAE decoder. The inpainting process is only guided by the text prompt at this time. When the lower part, including garment encoder and pose encoder, is added to the framework, the inpainting process can also be guided by the garment image. This figure shows that video virtual try-on is a conditional video inpainting task with the garment image as condition.}
	\label{fig:main_framework}
\end{figure*}

\subsection{Video Inpainting}
Video inpainting is the task of filling masked regions in a video, which is challenging due to its requirements of both spatial and temporal consistencies. AVID~\cite{zhang2024avid} is based on a text-to-image inpainting model and finetunes the motion module for temporal consistency, which works in a similar way as AnimateDiff~\cite{guo2023animatediff}. FGDVI~\cite{gu2023flow} utilizes a decoupled flow completion module to predict the flow from masked frames, and leverages optical flow as a guidance for diffusion model to improve inpainting quality and temporal consistency. In~\cite{green2024semantically}, they reframe video inpainting as a conditional generative modeling problem via a 4D-UNet and devise a method for conditioning on the known pixels in incomplete frames. Infusion~\cite{cherel2023infusion} also proposes a diffusion-based method, it adopts an internal learning approach, which leverages only the information contained in the current video. 
To the best of our knowledge, all the state-of-the-art diffusion-based video inpainting methods so far allow video input by inflating the U-Net with a temporal dimension. In this paper, we implement video inpainting with a full 3D attention diffusion transformer for better spatial-temporal consistency.

%% file: sec/2_formatting.tex
\section{Method}
\label{sec:formatting}

We formulate video virtual try-on as a conditional video inpainting task and we propose ViTI (Video Try-on Inpainter), a video virtual try-on framework built upon full 3D attention diffusion transformer, so in Sec.~\ref{sec_preliminaries} we give a brief introduction to diffuion model and diffusion transformer. Then in Sec.~\ref{sec_videogarmentinpainting} we introduce the video garment inpainting model, which is the base of ViTI. In Sec.~\ref{sec_videotryoninpainter} we add conditions to the video garment inpainting model to build ViTI.


\subsection{Preliminaries}
\label{sec_preliminaries}
\noindent\textbf{Diffusion model}. Latent diffusion model (LDM) conducts the diffusion process in the latent space instead of pixel space for efficiency. Given an input $x_0$, a task-specific encoder $\mathcal{E}$ is used to map it into the latent space $z_0=\mathcal{E}(x_0)$. During training, the diffusion process is applied to the latent $z_0$ in $t$ timesteps to produce a noisy latent $z_t$. And then the denoising diffusion model $\epsilon_{\theta}$ is trained to predict the added noise $\epsilon$ with the following loss function,

\begin{equation}
    \mathcal{L}_{LDM} = \mathbb{E}_{\epsilon\sim\mathcal{N}(0, 1),t,c}\lVert\epsilon - \epsilon_{\theta}(z_t, t, c)\rVert_2^2
\end{equation}
where $c$ denotes the conditional inputs, such as text prompt.

\noindent\textbf{Diffusion Transformer for Video Generation}. Different from U-Net, diffusion transformer (DiT) is based on vision transformer and it takes the token sequences instead of feature maps as input. In video generation, the input video is first compressed by a pretrained video VAE encoder to get latent feature maps, and then the feature maps are patchified and flatten into token sequences. Patchify module is essential in DiT because it is used to convert the spatial input into token sequences. In inference, the token sequence is first unpachified to spatial and then decoded with the video VAE decoder to pixel space. As for architecture, full 3D spatial-temporal attention directly calculates attention on the whole sequence, which has a strong spatial-temporal modeling ability but with a huge computation cost. Another popular architecture design follows Latte~\cite{ma2024latte}, which uses stacked spatial attention block and temporal attention block to approximate full attention, which is efficient but with poor spatial-temporal modeling ability.


\subsection{Video Garment Inpainting}
\label{sec_videogarmentinpainting}
\noindent\textbf{Overview}. In this section, a video inpainting framework with full 3D attention diffusion transformer is built at first, and then we progressively adapt it for video garment inpainting with the proposed masking strategies and multi-stage training.

\noindent\textbf{Video Inpainting with DiT}. Video inpainting is the task of filling masked regions in a video naturally according to the known regions and text guidance. Formally, given an $N$-frame video $x_0=\{x_0^i\}_{i=1}^N$, a corresponding $N$-frame binary mask $m_0=\{m_0^i\}_{i=1}^N$, and text prompt, the training objective for video inpainting is to minimize the reconstruction loss of the masked area,
\begin{equation}
    \mathcal{L}_{inpainting} = \sum_{i}\lVert (x_0^i - x_p^i) * \frac{m_0^i}{\sum{\mathbb{I}[m_0^i=1]}} \rVert_2^2
\end{equation}
where $x_p^i$ is the prediction for frame $i$, the values in $m_0^i$ equal to 1 if the pixels are to be modified, $\mathbb{I}[cond]$ equals to 1 if $cond$ is true otherwise 0.

In our implementation with DiT, there are slight differences because: 1) we use diffusion model so the loss is calculated on noises instead of pixel values; 2) we use a video VAE encoder $\mathcal{E}$ whose compression rate in three axis are all larger than 1, so the operations are done in latent space with a smaller spatial and temporal size; 3) the original mask $m_0$ should be transformed to latent space with a reshape operator $\mathcal{R}$ to match token shapes. Let $z_0=\mathcal{E}(x_0)=\{z_0^i\}_{i=1}^T$, $m_z=\mathcal{R}(m_0)=\{m_z^i\}_{i=1}^T$, the diffusion transformer for video inpainting is trained with the following loss,

\begin{equation}
    \mathcal{L} = \mathbb{E}_{\epsilon\sim\mathcal{N}(0, 1),t,c}\lVert(\epsilon - \epsilon_{\theta}(z_t, t, c)) * \frac{m_z}{\sum\mathbb{I}[m_z \neq 0]}\rVert_2^2
\end{equation}
where $z_t$ is a noisy latent which is diffused from $z_0$ with $t$ timesteps.

\begin{figure}[t]
	\centering
	\includegraphics[scale=0.21]{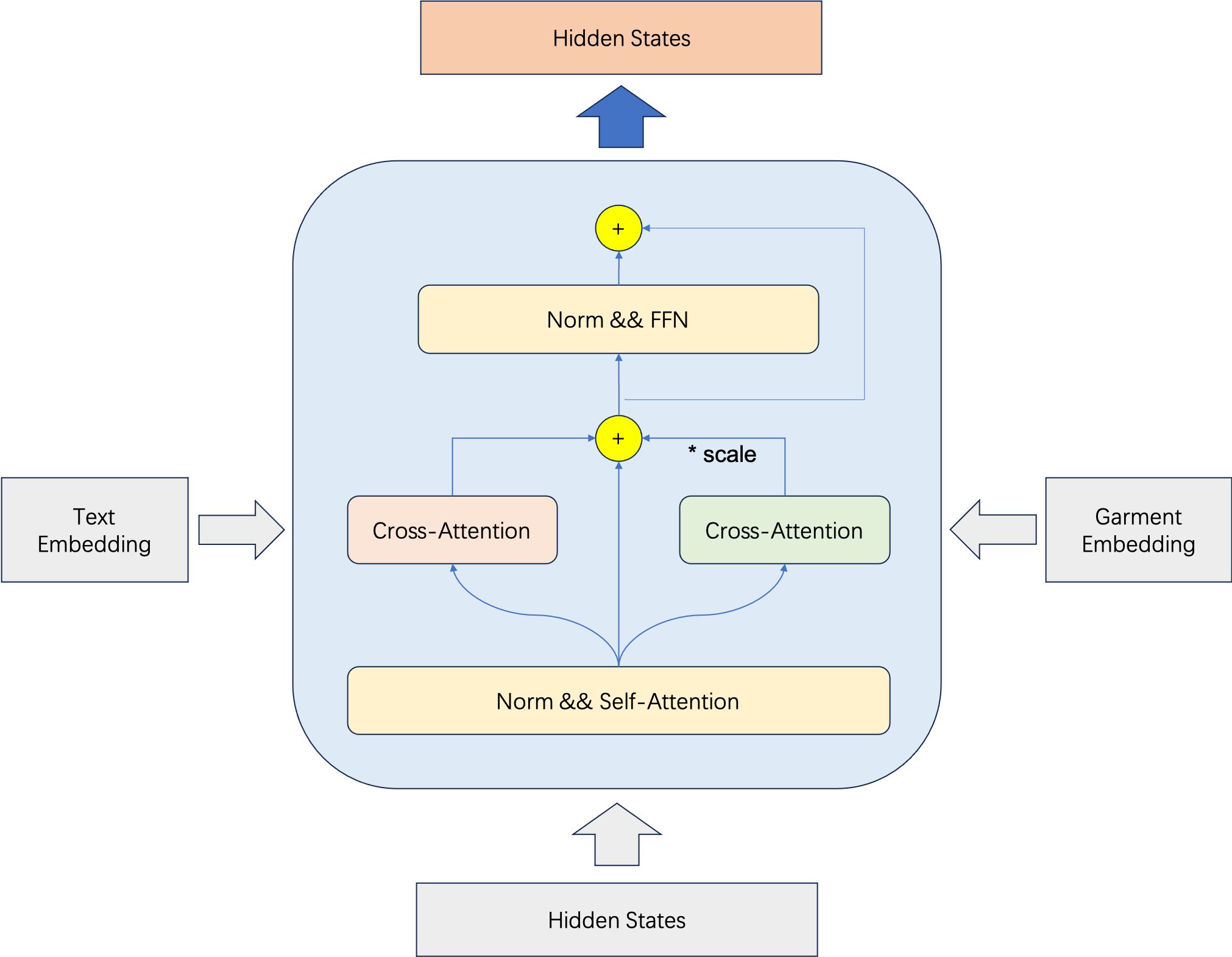}
	\caption{Full 3D attention transformer layer. The garment embeddings interact with the block through an additional cross-attention, which lies in parallel to the text embedding cross-attention, with a scale factor to control the guidance strength.}
	\label{fig:attention_layer}
\end{figure}

\noindent\textbf{Network Architecture}. The proposed DiT based video inpainting framework is illustrated in Figure.~\ref{fig:main_framework}. It consists of four parts, a text encoder, a mask reshaper, a video VAE and a diffusion transformer. The text encoder is used to extract text embeddings from user input prompt, which acts as the generation guidance for video inpainting, we choose T5~\cite{raffel2023exploringlimitstransferlearning} as the text encoder because it offers good prompt following ability in generation tasks. The reshaper projects the original mask from pixel space to latent space, which is mainly used for token shape conversion, and it is implemented by interpolation operator. The video VAE encoder is used to compress video, which can significantly reduce the token numbers, and video VAE decoder recovers the video from latent space to pixel space, we use causal video VAE~\cite{pku_yuan_lab_and_tuzhan_ai_etc_2024_10948109} because it provides a temporal compression, which can further reduce the length of token sequence.

As the core of the framework, the diffusion transformer is composed of stacked full 3D attention transformer layers. As illustrated in Figure.~\ref{fig:attention_layer}, the full 3D attention block consists of pre-normalization, self-attention, cross-attention, residual connection, post-normalization and FFN. For video inpainting, only the cross-attention with text embedding is essential. The input hidden states comes from the patchified and flattened token sequence, and it is feed into full 3D spatial-temporal attention to update itself.

It is noteworthy that the input noise to DiT is the addition of three parts: reshaped mask $m_z$, diffused latent $z_t$, and masked video latent $\mathcal{E}(x_0 \bigodot (1-m_0))$, where $\bigodot$ is pixel-wise multiplication.

\noindent\textbf{Training strategy}. 
To train the video garment inpainting model, we carefully design a collection of masking strategies, and then we propose multi-stage training to progressively adapt a pretrained text-to-video model to a video garment inpainting model. The core reason for us to design such multi-stage training with different masking strategies is the lack of high quality data, because we have too little videos with precise garment masks to train the garment inpainting model all at once.

\begin{figure}[t]
	\centering
	\includegraphics[scale=0.20]{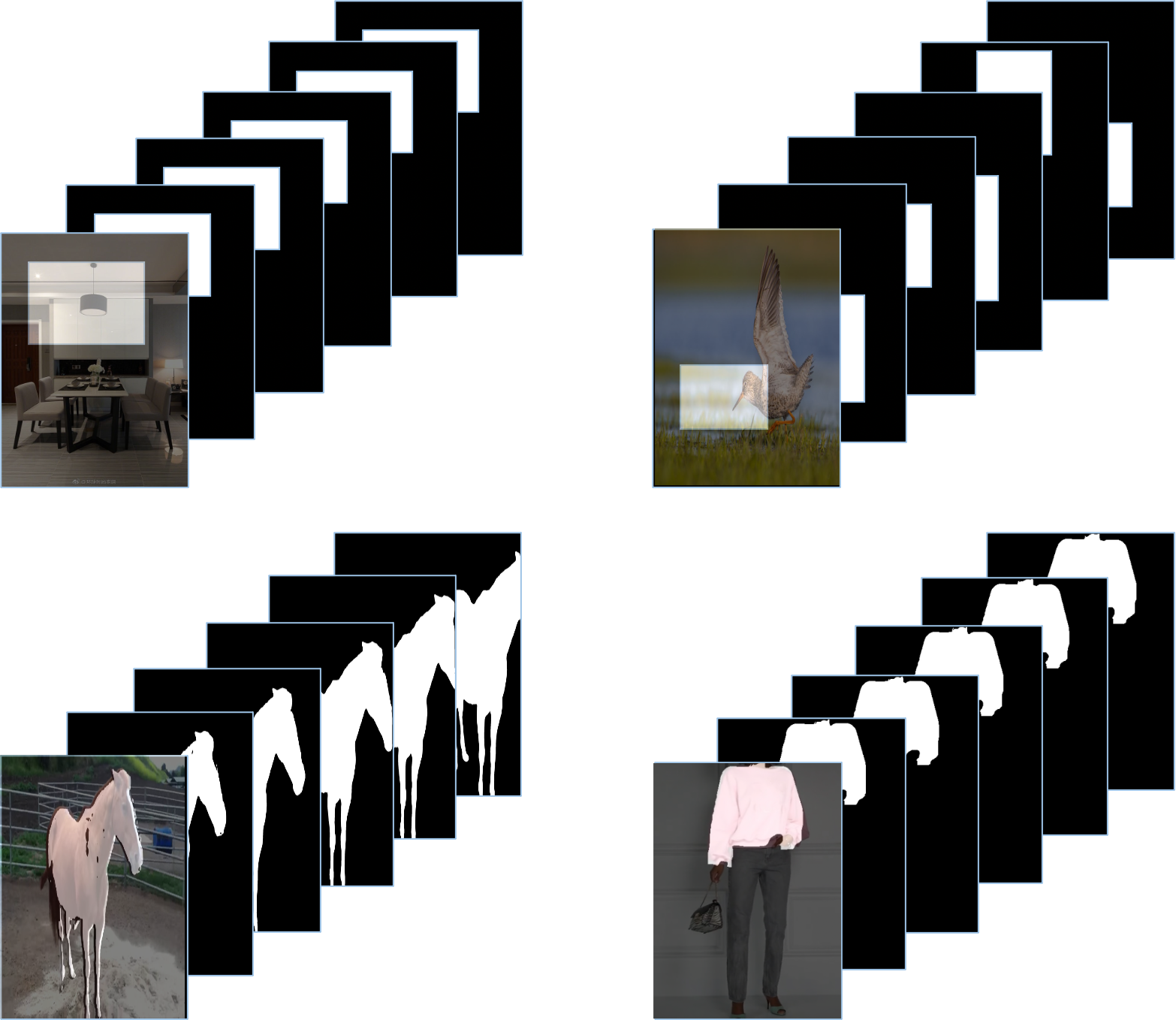}
	\caption{Masking strategies for garment inpainting model. These are time-invariant box mask, time-variant box mask, instance-level mask and garment mask.}
	\label{fig:masking_strategy}
\end{figure}

As illustrated in Figure.~\ref{fig:masking_strategy}, we design four kind of masking strategies, time-invariant box mask, time-variant box mask, instance-level mask and garment mask. In time-invariant box mask, the position and size of the mask is random, but all the frames share the same mask, while in time-variant box mask, the position and size of the mask can be different between frames. These two kind of masks are general and can be applied to any videos, so they are used in the early training stage with a large video dataset~\cite{yang2019videoinstancesegmentation,ding2023mevislargescalebenchmarkvideo}. And during training, there is a probability to inverse these masks to improve the diversity. As for instance-level mask, the mask in each frame is a segmentation mask of the content of interest, and similar in garment mask, the mask in each frame is a segmentation mask of the garment. These two kind of masks are used in the later training stage, because only few videos have precise instance-level mask and garment mask due to the high acquisition cost of such data.

Then we train the video garment inpainting model in multiple stages with different dataset. Specifically, in the first stage (Stage 1), the model is trained on a large video dataset~\cite{ding2023mevislargescalebenchmarkvideo,yang2019videoinstancesegmentation}, with time-invariant box mask and time-variant box mask strategies. The text prompt for each video is the original caption from the video dataset, though not semantically aligned with the mask area, still works in practice. After this stage, the model has a basic video inpainting ability given coarse masks. In the second stage (Stage 2), the model is trained on VOS (Video Object Segmentation) dataset~\cite{Caelles_arXiv_2019,ponttuset20182017davischallengevideo,hong2024lvosbenchmarklargescalelongterm}, with instance-level mask. The text prompt for each video comes from the `motion expression' (indeed also image captions) of the target object, which has a good alignment with the mask area. In the third stage (Stage 3), the model is trained on our collected human-centric dataset VTP, with garment mask, and the text prompt is associated with the garment mask area. The collection of the dataset VTP is detailed in experiments part. 

After these training stages, the model can inpaint the masked garment area with garment pixels according to the text prompt with good spatial-temporal consistency. In the next section, we introduce conditions to this model to fullfill the task of video virtual try-on, which can control the appearance and details of the garment by reference image.

\begin{figure}
    \centering

    \includegraphics[scale=0.14]{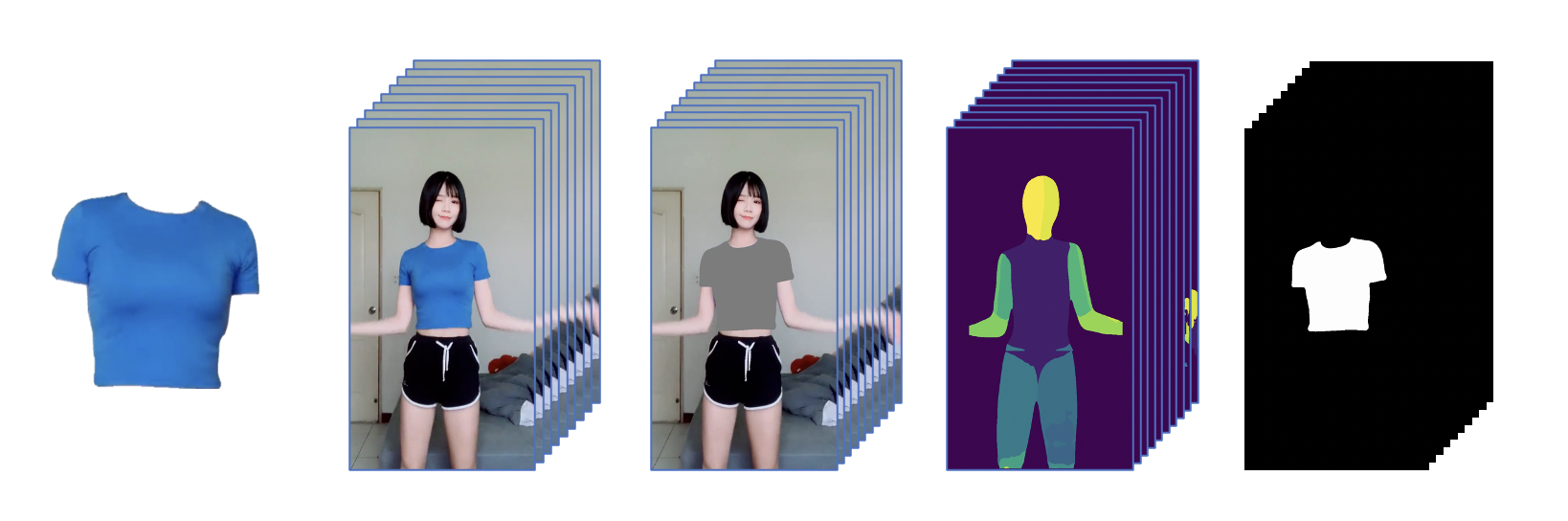}
    \caption{Visualization of an example of the proposed dataset VTP for video try-on pretraining. From left to right, these are garment image, original video, garment-agnostic video, densepose video and garment mask video. It is noteworthy that there is not a standard garment image for each Internet-sourced video, so the shown garment image (leftmost) is cropped from one of the original video frames via its corresponding garment mask.}
    \label{fig:dataset_visualization}
\end{figure}

\subsection{Video Try-on Inpainter (ViTI)}
\label{sec_videotryoninpainter}
\noindent\textbf{Overview}. Based on the above video inpainting framework, conditions are added in this section to make video try-on a conditional video inpainting task. As illustrated in the lower part of Figure.~\ref{fig:main_framework}, a garment encoder on the right is added to extract garment embedding as condition to the DiT block through cross-attention. A pose encoder on the left is added to extract pose latent as content prior. Also, a loss to constrain the difference between consecutive frames is explicitly applied to further improve the temporal consistency.

\noindent\textbf{Garment encoder}. One of the key problems for virtual try-on is garment detail preservation. We design garment encoder to extract the visual feature of the reference garment image. It mainly consists of two parts, an image VAE encoder~\cite{zhu2023designingbetterasymmetricvqgan} and a DINOv2~\cite{oquab2024dinov2learningrobustvisual}. The image VAE encoded feature and the DINOv2 extracted feature, are first transformed by linear layers, and then concatenated together to send into an MLP layer to produce the output garment embedding. DINOv2 is used because its visual features are robust and perform well across domains, which meets the need of extracting robust feature for the garment image.

\noindent\textbf{Pose encoder}. Pose encoder is composed of a DensePose~\cite{güler2018denseposedensehumanpose} encoder followed by an MLP layer, the output of which has the same tensor shape with latent noise, and it is fused with other latents as input to the DiT model through addition in latent space. DensePose is used because we hypothesize that the depth information in it can provide consistent spatial content prior.

\noindent\textbf{Garment adapter}. The garment embedding is injected into full 3D attention transformer layer by cross-attention, a similar way as in IPAdapter~\cite{ye2023ip-adapter}. As illustrated in Figure.~\ref{fig:attention_layer}, the cross-attention for garment embedding lies in parallel to the text embedding cross-attention, with a scale factor to control the guidance strength.

\noindent\textbf{Temporal consistency loss}. Although full 3D attention diffusion transformer has a strong spatial-temporal modeling ability, there still exist flickers in the generated videos because of the limited high quality training data. Thus we seek to explicitly improve the temporal consistency. We use the following loss to constrain the difference between consecutive latent frames,

\begin{equation}
    \mathcal{L}_{temporal} = \sum_{i=1}^{T-1} \lVert\epsilon_{\theta}(z_t, t, c)^i - \epsilon_{\theta}(z_t, t, c)^{i+1}\rVert_2^2
\end{equation}
where $\epsilon_{\theta}(*)^i$ is the sub-latent in the $i$-th frame. So the total loss for training ViTI is as follows,
\begin{equation}
    \mathcal{L}_{total} = \mathcal{L} + \alpha * \mathcal{L}_{temporal}
\end{equation}
where $\alpha=0.1$ is a hyperparameter. Initialized with the weights of the pretrained video garment inpainting model, ViTI is optimized with $\mathcal{L}_{total}$ on dataset Vivid~\cite{fang2024vivid}.

\section{Experiments}

In this section, we first give an introduction 
to the dataset,
then we compare the proposed ViTI with other state-of-the-art methods both qualitatively and quantitatively, and finally we conduct ablation studies to illustrate some of the key design choices behind ViTI.

\subsection{Implementation Details}

\noindent\textbf{Dataset preparation}. For video virtual try-on task, there are only two publicly available dataset, VVT~\cite{jiang2022clothformer} and Vivid~\cite{fang2024vivid}. VVT is widely used for benchmarking, it includes 791 videos with corresponding garment images at a resolution of 256 × 192. Vivid comprises 9,700 clothing-video pairs with a resolution of 832 × 624, categorized into three clothing types, upper-body, lower-body, and dresses. These datasets with high quality are more suitable for finetuning, which are usually used in the last training stage. However, as we know, data plays a key role in two aspects, the quality and the amount. Vivid provides a small amount of high quality dataset, while no large amount dataset for video try-on pretraining. So we construct a human-centric video dataset for Video Try-on Pretraining (VTP) with a variety of clothing styles, body movements and backgrounds, which is specifically designed for early stage pretraining. Using a collection of tools, such as human detection, parsing, blur evaluation and aesthetic evaluation, we obtain a total of 51,278 video clips in the end, each of which contains a single human with enough upper body or lower body ratio, Figure.~\ref{fig:dataset_visualization} shows an example. For more information about VTP please refer to the supplementary materials.

\noindent\textbf{Experimental Settings}. The model is trained across 8 Nvidia A100 GPUs using the AdamW optimizer with a learning rate of 1e-5. The resolution of the video clip for training is set to 29 x 512 × 384.

\begin{table}[t]
\centering
\begin{tabular}{lcccc}
\toprule
\textbf{Method} & \textbf{SSIM $\uparrow$} & \textbf{LPIPS $\downarrow$} & \textbf{VFID(I3D) $\downarrow$} \\
\midrule
FW\_GAN~\cite{dong2019fw}          & 0.675 & 0.283 & 8.019  \\
CP-VTON~\cite{wang2018characteristicpreservingimagebasedvirtualtryon}          & 0.459 & 0.535 & 6.361  \\
PBAFN~\cite{ge2021parserfreevirtualtryondistilling}             & 0.870 & 0.157 & 4.516  \\
WildVidFit~\cite{he2024wildvidfit}      & - & - & 4.202  \\
OOTDiffusion~\cite{xu2024ootdiffusion}      & 0.882 & 0.070 & 4.167  \\
StableVITON~\cite{kim2023stableviton}      & 0.876 & 0.076 & 4.021  \\
ClothFormer~\cite{jiang2022clothformer}       & 0.921 & 0.081 & 3.967  \\
VIVID~\cite{fang2024vivid}            & \textbf{0.949} & 0.068 & 3.405  \\
Tunnel Try-on~\cite{xu2024tunnel}      & 0.913 & 0.054 & 3.345  \\
ViTI (Ours)         & 0.938 & \textbf{0.042} & \textbf{2.121}  \\
\bottomrule
\end{tabular}
\caption{Quantitative comparisons on VVT dataset. The best results are highlighted in bold. As can be seen, ViTI achieves significant improvements on VFID. Result of WildVidFit~\cite{he2024wildvidfit} is partially reported due to the lack of open-sourced model. A whole table with VFID (Resnext3d) please refer to the supplementaries.}
\label{tab_all}
\end{table}

\subsection{ Qualitative Results}
\begin{figure}[t]
	\centering
	\includegraphics[scale=0.80]{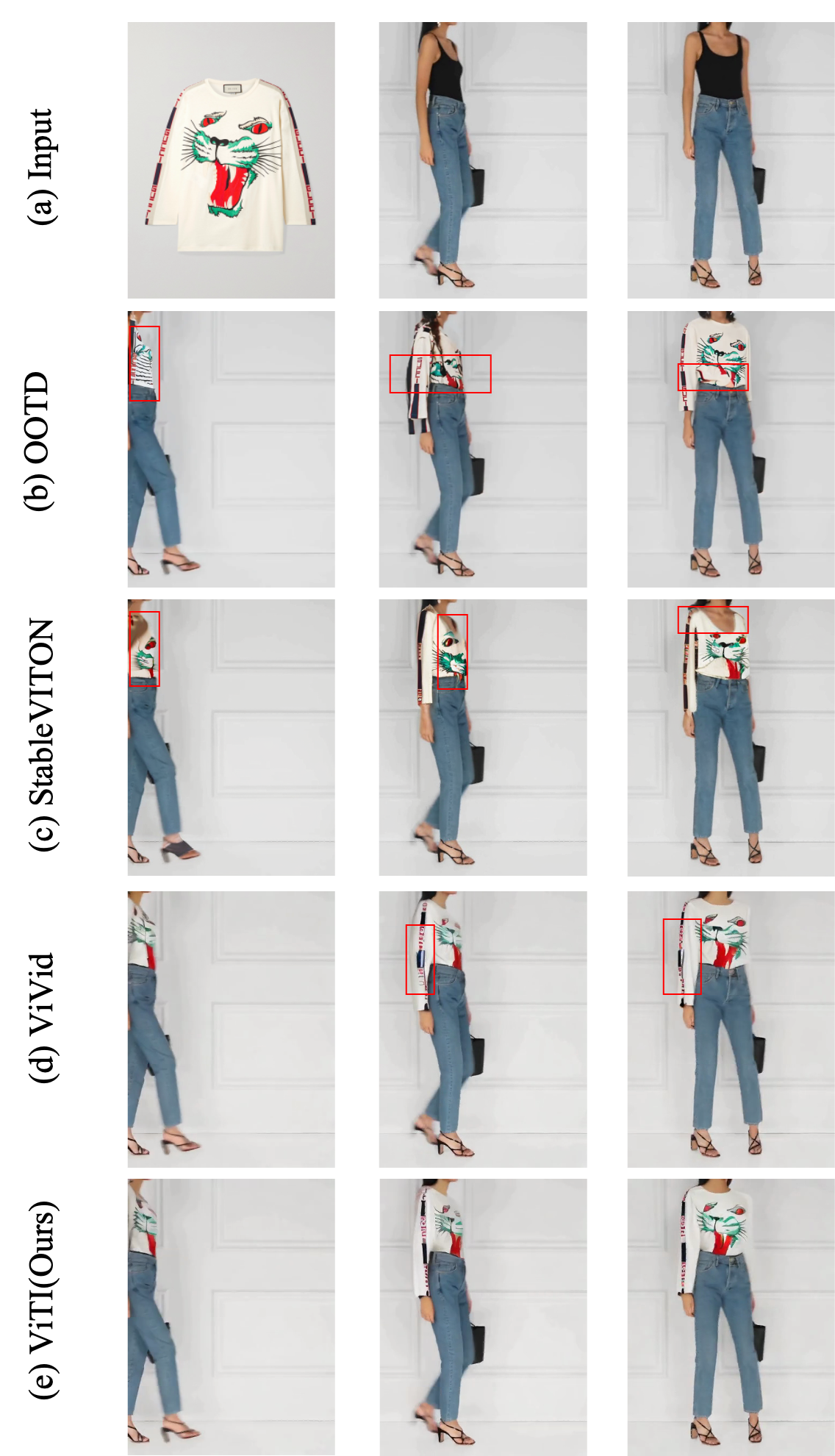}
	\caption{Qualitative comparisons with other methods on the public dataset. ViTI achieves high-detail preservation and good temporal consistency, and performs better than any other methods.}
	\label{fig:compare results}
\end{figure}

We conduct a qualitative comparison with several recent state-of-the-art image-based try-on methods (OOTDiffusion~\cite{xu2024ootdiffusion} and StableVTON~\cite{kim2024stableviton}) and diffusion-based video try-on method (Vivid~\cite{fang2024vivid}). Figure.~\ref{fig:compare results} shows the visual comparison result, as can be seen, our method ViTI obviously outperforms other methods. The image-based try-on methods, on the one hand, struggle with handling side-view scenes effectively, and the detail of the garment is also not satisfactory. On the other hand, due to the lack of temporal information, they fail to achieve temporal consistency. The diffusion-based state-of-the-art method Vivid, which incorporates a temporal module, shows notable improvement in temporal consistency, but still lacks of sufficient preservation of garment details. Figure.\ref{fig:first_image}  showcases a range of results generated by ViTI, encompassing various scenes and garment types. Our method consistently achieves high-detail preservation and maintains temporal consistency across the generated try-on videos. For more visualization results, please refer to the supplementaries.

\subsection{Quantitative Results}

\noindent\textbf{Evaluation Metrics}. For video try-on evaluation, Video Frechet Inception Distance (VFID)~\cite{unterthiner2018towards} is used to measure both visual quality and temporal consistency of the generated results. There are two kind of 3D convolution feature extractor used for VFID, one is I3D~\cite{carreira2018quovadisactionrecognition} and the other is 3D-ResNeXt101~\cite{hara2018spatiotemporal3dcnnsretrace}. For the single-frame evaluation, we rely on two primary metrics: Structural Similarity Index (SSIM)~\cite{wang2004image} and Learned Perceptual Image Patch Similarity (LPIPS)~\cite{zhang2018unreasonableeffectivenessdeepfeatures}. These metrics are used to assess the quality of individual frames in a paired comparison setting. A higher SSIM value or a lower LPIPS score indicates better visual similarity between the generated frame and the original one. Although SSIM and LPIPS are not sufficient to evaluate the quality of videos, we still use them for video frame quality evaluation as a supplement.


We perform a quantitative evaluation on the dataset VVT~\cite{jiang2022clothformer} with existing visual try-on methods,
including GAN based methods like FW-GAN~\cite{dong2019fw}, PBAFN~\cite{Yuying2021Parser-free} and Cloth Former~\cite{jiang2022clothformer}, and diffusion-based methods like Vivid~\cite{fang2024vivid}. Some image-based try-on algorithms are also evaluated, including CP-VTON~\cite{wang2018toward} and StableVITON~\cite{kim2024stableviton}. As shown in Table.~\ref{tab_all}, the proposed method ViTI achieves state-of-the-art performance on most metrics, especially on VFID with a significant improvement compared to previous methods.


\subsection{ Ablation Study }
In this section, we conduct ablation experiments to illustrate some key design choices behind ViTI.


\noindent\textbf{Garment encoder}. Detail preservation is important to try-on task, so image encoders those can capture detail information are preferred. VAE is trained to compress the image and then recover it from the compressed latent, so VAE is a first choice because its latent contains much detail information used for reconstruction. However using image VAE encoder alone yields poor results, mainly due to the lack of semantic information. So we introduce CLIP vision encoder for high-level features and combine the VAE encoder and CLIP vision encoder together, which achieves better result. To put it further, we replace CLIP vision encoder with DINOv2, a more robust vision encoder, which focuses on self-supervised pre-training with extensive collections of clustered images, can produce features with strong semantic as well as much details. Table.~\ref{tab_garmentencoder} reports the experiment results of different garment encoder choices.

\begin{table}[t]
\centering
\begin{tabular}{lcccc}
\toprule
\textbf{Method} & \textbf{SSIM $\uparrow$} & \textbf{LPIPS $\downarrow$} & \textbf{VFID(I3D) $\downarrow$}  \\
\midrule

VAE     & 0.889 & 0.072  &  3.421 \\
VAE+CLIP             & 0.921 & 0.059 & 2.356  \\
VAE+DINOv2        & \textbf{0.938} & \textbf{0.042} & \textbf{2.121}  \\
\bottomrule
\end{tabular}
\caption{Ablation study on the garment encoder. }
\label{tab_garmentencoder}
\end{table} 

\begin{table}[t]
\centering
\begin{tabular}{cccccc}
\toprule
 \textbf{PoseE} & \textbf{TCLoss}  & \textbf{SSIM}$\uparrow$  &   \textbf{LPIPS}$\downarrow$ & \textbf{VFID(I3D) $\downarrow$}  \\
\midrule
  & & 0.911 & 0.062  & 2.446 \\
 \checkmark & & 0.929 & 0.049&   2.223 \\
  &\checkmark  & 0.920 & 0.054 &  2.359\\
 \checkmark & \checkmark &\textbf{0.938}  & \textbf{0.042}  &\textbf{2.121} \\
\bottomrule
\end{tabular}
\caption{Ablation study on the pose encoder (PoseE) and the temporal consistency loss (TCLoss).}
\label{tab_densepose_tcloss}
\end{table}

\begin{table}[t]
\centering
\begin{tabular}{lcccc}
\toprule
\textbf{Method} & \textbf{SSIM $\uparrow$} & \textbf{LPIPS $\downarrow$} & \textbf{VFID(I3D) $\downarrow$} \\
\midrule
ViTI (w/o VTP) & 0.922 & 0.055 & 2.935 \\
ViTI    & \textbf{0.938} & \textbf{0.042} & \textbf{2.121} \\
\bottomrule
\end{tabular}
\caption{Ablation study on the VTP dataset. It's noteworthy that even without VTP dataset, ViTI achieves best VFID(I3D) compared to other methods as shown in Table~\ref{tab_all}.}
\label{tab_vtp}
\end{table}

\begin{table}[t]
\centering
\begin{tabular}{lcccc}
\toprule
\textbf{Method} & \textbf{Stage 1} & \textbf{Stage 2} & \textbf{Stage 3} \\
\midrule
Imaging Quality $\uparrow$ & 0.3940 & 0.4271 & \textbf{0.4303} \\
Subject Consistency $\uparrow$ & 0.4111 & 0.4129 & \textbf{0.4171} \\
Temporal Flickering $\uparrow$ & 0.9801 & 0.9814 & \textbf{0.9821} \\
Motion Smoothness $\uparrow$ & 0.9930 & \textbf{0.9937} & 0.9936 \\
Overall Consistency $\uparrow$ & 0.1387 & 0.1659 & \textbf{0.2096} \\
\bottomrule
\end{tabular}
\caption{Quantitative analysis of different stages. Because there is no benchmark to evaluate the garment inpainting task, so we use metrics from VBench to evaluate the video quality.}
\label{tab_multi_stages}
\end{table}

\noindent\textbf{Pose encoder}. Densepose information is considered to be a valuable conditional input for capturing depth information, which is often critical for non-rigid objects, so DensePose is introduced as pose encoder to deal with those non-rigid objects (garment and human body) in try-on task for better spatial-temporal consistency, in the hypothesis that depth information is continuous in the spatial-temporal space. Table.~\ref{tab_densepose_tcloss} reports the experiment results of using or not using pose encoder. It can be seen that using pose encoder shows better performance.



\noindent\textbf{Temporal consistency loss}. Similarly, we evaluate the effectiveness of the temporal consistency loss by using it or not using it during training. Table.~\ref{tab_densepose_tcloss} reports the experiment results, it can be seen that there is a performance improvement by using temporal consistency loss. So we can explicitly constrain the difference between consecutive latent frames to get better temporal consistency, especially when the amount of high quality training data is small.


\noindent\textbf{Multi-stage training}. As mentioned in Sec.~\ref{sec_videogarmentinpainting}, the core reason to design multi-stage training with different masking strategies
is the lack of high quality data, because we have too little videos with precise garment masks to train the garment inpainting model all at once. So we have to progressively make it. It's noteworthy that after Stage 1/2/3 training, we only get a video garment inpainting model, which can not be evaluated on VVT, and also there is no specific benchmark to evaluate this kind of task, so we use metrics from VBench to evaluate the video quality as a reference. Table.~\ref{tab_multi_stages} reports the quantitative results after different stage training. As can be seen, the performance improves steady via progressive training. Typical visualization results of different stages please refer to supplementary materials.

\noindent\textbf{Video try-on pretraining (VTP) dataset}. With only public video try-on dataset (such as Vivid) we find that although the model has the ability to try-on clothes in most cases, it shows poor robustness when doing out-of-distribution test, and it often has unexpected flickering, implying that the model is not sufficiently trained. So we collect VTP, a dataset for video try-on pretraining, which is larger than the previous video try-on dataset. With VTP, ViTI shows not only better but also robust performance. Table.~\ref{tab_vtp} shows the effectiveness of the VTP dataset.





%% file: sec/3_finalcopy.tex
\section{Conclusions}

We propose ViTI and formulate video try-on as a conditional video inpainting task. Specifically, we first build a video inpainting framework based on diffsion transformer with full 3D spatial-temporal attention. And then we progressively adapt it for video garment inpainting with a collection of masking strategies and multi-stage training. And to train the model, we collect a human-centric dataset with 51,278 video clips for video try-on pretraining. 
Finally, garment condition is added to make sure the inpainted garment appearance and details are as expected. ViTI utilizes a completely different framework to deal with video try-on task compared to previous methods, which shows significant improvements. We hope our method can
bring new insight to the community.

%% file: main.bbl
\begin{thebibliography}{46}
\providecommand{\natexlab}[1]{#1}
\providecommand{\url}[1]{\texttt{#1}}
\expandafter\ifx\csname urlstyle\endcsname\relax
  \providecommand{\doi}[1]{doi: #1}\else
  \providecommand{\doi}{doi: \begingroup \urlstyle{rm}\Url}\fi

\bibitem[Baldrati et~al.(2023)Baldrati, Morelli, Cartella, Cornia, Bertini, and Cucchiara]{baldrati2023multimodal}
Alberto Baldrati, Davide Morelli, Giuseppe Cartella, Marcella Cornia, Marco Bertini, and Rita Cucchiara.
\newblock Multimodal garment designer: Human-centric latent diffusion models for fashion image editing.
\newblock In \emph{Proceedings of the IEEE/CVF International Conference on Computer Vision}, pages 23393--23402, 2023.

\bibitem[Caelles et~al.(2019)Caelles, Pont-Tuset, Perazzi, Montes, Maninis, and {Van Gool}]{Caelles_arXiv_2019}
Sergi Caelles, Jordi Pont-Tuset, Federico Perazzi, Alberto Montes, Kevis-Kokitsi Maninis, and Luc {Van Gool}.
\newblock The 2019 davis challenge on vos: Unsupervised multi-object segmentation.
\newblock \emph{arXiv:1905.00737}, 2019.

\bibitem[Carreira and Zisserman(2018)]{carreira2018quovadisactionrecognition}
Joao Carreira and Andrew Zisserman.
\newblock Quo vadis, action recognition? a new model and the kinetics dataset, 2018.

\bibitem[Cherel et~al.(2023)Cherel, Almansa, Gousseau, and Newson]{cherel2023infusion}
Nicolas Cherel, Andr{\'e}s Almansa, Yann Gousseau, and Alasdair Newson.
\newblock Infusion: Internal diffusion for video inpainting.
\newblock \emph{arXiv preprint arXiv:2311.01090}, 2023.

\bibitem[Choi et~al.(2021)Choi, Park, Lee, and Choo]{choi2021viton}
Seunghwan Choi, Sunghyun Park, Minsoo Lee, and Jaegul Choo.
\newblock Viton-hd: High-resolution virtual try-on via misalignment-aware normalization.
\newblock In \emph{Proc. of the IEEE conference on computer vision and pattern recognition (CVPR)}, 2021.

\bibitem[Ding et~al.(2023)Ding, Liu, He, Jiang, and Loy]{ding2023mevislargescalebenchmarkvideo}
Henghui Ding, Chang Liu, Shuting He, Xudong Jiang, and Chen~Change Loy.
\newblock Mevis: A large-scale benchmark for video segmentation with motion expressions, 2023.

\bibitem[Dong et~al.(2019)Dong, Liang, Shen, Wu, Chen, and Yin]{dong2019fw}
Haoye Dong, Xiaodan Liang, Xiaohui Shen, Bowen Wu, Bing-Cheng Chen, and Jian Yin.
\newblock Fw-gan: Flow-navigated warping gan for video virtual try-on.
\newblock In \emph{Proceedings of the IEEE/CVF international conference on computer vision}, pages 1161--1170, 2019.

\bibitem[Fang et~al.(2024)Fang, Zhai, Su, Song, Zhu, Wang, Chen, Liu, Cao, and Zha]{fang2024vivid}
Zixun Fang, Wei Zhai, Aimin Su, Hongliang Song, Kai Zhu, Mao Wang, Yu Chen, Zhiheng Liu, Yang Cao, and Zheng-Jun Zha.
\newblock Vivid: Video virtual try-on using diffusion models.
\newblock 2024.

\bibitem[Ge et~al.(2021{\natexlab{a}})Ge, Song, Zhang, Ge, Liu, and Luo]{Yuying2021Parser-free}
Yuying Ge, Yibing Song, Ruimao Zhang, Chongjian Ge, Wei Liu, and Ping Luo.
\newblock Parser-free virtual try-on via distilling appearance flows.
\newblock In \emph{In Proceedings of the IEEE/CVF Conference on Computer Vision and Pattern Recognition}, page 8485–8493, 2021{\natexlab{a}}.

\bibitem[Ge et~al.(2021{\natexlab{b}})Ge, Song, Zhang, Ge, Liu, and Luo]{ge2021parser}
Yuying Ge, Yibing Song, Ruimao Zhang, Chongjian Ge, Wei Liu, and Ping Luo.
\newblock Parser-free virtual try-on via distilling appearance flows.
\newblock In \emph{Proceedings of the IEEE/CVF conference on computer vision and pattern recognition}, pages 8485--8493, 2021{\natexlab{b}}.

\bibitem[Ge et~al.(2021{\natexlab{c}})Ge, Song, Zhang, Ge, Liu, and Luo]{ge2021parserfreevirtualtryondistilling}
Yuying Ge, Yibing Song, Ruimao Zhang, Chongjian Ge, Wei Liu, and Ping Luo.
\newblock Parser-free virtual try-on via distilling appearance flows, 2021{\natexlab{c}}.

\bibitem[Gou et~al.(2023)Gou, Sun, Zhang, Si, Qian, and Zhang]{gou2023taming}
Junhong Gou, Siyu Sun, Jianfu Zhang, Jianlou Si, Chen Qian, and Liqing Zhang.
\newblock Taming the power of diffusion models for high-quality virtual try-on with appearance flow.
\newblock \emph{arXiv preprint arXiv:2308.06101}, 2023.

\bibitem[Green et~al.(2024)Green, Harvey, Naderiparizi, Niedoba, Liu, Liang, Lavington, Zhang, Lioutas, Dabiri, et~al.]{green2024semantically}
Dylan Green, William Harvey, Saeid Naderiparizi, Matthew Niedoba, Yunpeng Liu, Xiaoxuan Liang, Jonathan Lavington, Ke Zhang, Vasileios Lioutas, Setareh Dabiri, et~al.
\newblock Semantically consistent video inpainting with conditional diffusion models.
\newblock \emph{arXiv preprint arXiv:2405.00251}, 2024.

\bibitem[Gu et~al.(2023)Gu, Yu, Fan, and Zhang]{gu2023flow}
Bohai Gu, Yongsheng Yu, Heng Fan, and Libo Zhang.
\newblock Flow-guided diffusion for video inpainting.
\newblock \emph{arXiv preprint arXiv:2311.15368}, 2023.

\bibitem[Guo et~al.(2024)Guo, Yang, Rao, Liang, Wang, Qiao, Agrawala, Lin, and Dai]{guo2023animatediff}
Yuwei Guo, Ceyuan Yang, Anyi Rao, Zhengyang Liang, Yaohui Wang, Yu Qiao, Maneesh Agrawala, Dahua Lin, and Bo Dai.
\newblock Animatediff: Animate your personalized text-to-image diffusion models without specific tuning.
\newblock \emph{International Conference on Learning Representations}, 2024.

\bibitem[Güler et~al.(2018)Güler, Neverova, and Kokkinos]{güler2018denseposedensehumanpose}
Rıza~Alp Güler, Natalia Neverova, and Iasonas Kokkinos.
\newblock Densepose: Dense human pose estimation in the wild, 2018.

\bibitem[Han et~al.(2018)Han, Wu, Wu, Yu, and Davis]{han2017viton}
Xintong Han, Zuxuan Wu, Zhe Wu, Ruichi Yu, and Larry~S Davis.
\newblock Viton: An image-based virtual try-on network.
\newblock In \emph{CVPR}, 2018.

\bibitem[Hara et~al.(2018)Hara, Kataoka, and Satoh]{hara2018spatiotemporal3dcnnsretrace}
Kensho Hara, Hirokatsu Kataoka, and Yutaka Satoh.
\newblock Can spatiotemporal 3d cnns retrace the history of 2d cnns and imagenet?, 2018.

\bibitem[He et~al.(2024)He, Chen, Wang, Li, Torr, and Lin]{he2024wildvidfit}
Zijian He, Peixin Chen, Guangrun Wang, Guanbin Li, Philip~HS Torr, and Liang Lin.
\newblock Wildvidfit: Video virtual try-on in the wild via image-based controlled diffusion models.
\newblock \emph{arXiv preprint arXiv:2407.10625}, 2024.

\bibitem[Hong et~al.(2024)Hong, Liu, Chen, Tan, Feng, Zhou, Guo, Li, Chen, Gao, Zhang, and Zhang]{hong2024lvosbenchmarklargescalelongterm}
Lingyi Hong, Zhongying Liu, Wenchao Chen, Chenzhi Tan, Yuang Feng, Xinyu Zhou, Pinxue Guo, Jinglun Li, Zhaoyu Chen, Shuyong Gao, Wei Zhang, and Wenqiang Zhang.
\newblock Lvos: A benchmark for large-scale long-term video object segmentation, 2024.

\bibitem[Issenhuth et~al.(2020)Issenhuth, Mary, and Calauzenes]{issenhuth2020not}
Thibaut Issenhuth, J{\'e}r{\'e}mie Mary, and Cl{\'e}ment Calauzenes.
\newblock Do not mask what you do not need to mask: a parser-free virtual try-on.
\newblock In \emph{Computer Vision--ECCV 2020: 16th European Conference, Glasgow, UK, August 23--28, 2020, Proceedings, Part XX 16}, pages 619--635. Springer, 2020.

\bibitem[Jiang et~al.(2022)Jiang, Wang, Yan, and Liu]{jiang2022clothformer}
Jianbin Jiang, Tan Wang, He Yan, and Junhui Liu.
\newblock Clothformer: Taming video virtual try-on in all module.
\newblock In \emph{Proceedings of the IEEE Conference on Computer Vision and Pattern Recognition (CVPR)}, 2022.

\bibitem[Khirodkar et~al.(2024)Khirodkar, Bagautdinov, Martinez, Zhaoen, James, Selednik, Anderson, and Saito]{khirodkar2024sapiens}
Rawal Khirodkar, Timur Bagautdinov, Julieta Martinez, Su Zhaoen, Austin James, Peter Selednik, Stuart Anderson, and Shunsuke Saito.
\newblock Sapiens: Foundation for human vision models.
\newblock \emph{arXiv preprint arXiv:2408.12569}, 2024.

\bibitem[Kim et~al.(2023)Kim, Gu, Park, Park, and Choo]{kim2023stableviton}
Jeongho Kim, Gyojung Gu, Minho Park, Sunghyun Park, and Jaegul Choo.
\newblock Stableviton: Learning semantic correspondence with latent diffusion model for virtual try-on.
\newblock \emph{arXiv preprint arXiv:2312.01725}, 2023.

\bibitem[Kim et~al.(2024)Kim, Gu, Park, Park, and Choo]{kim2024stableviton}
Jeongho Kim, Guojung Gu, Minho Park, Sunghyun Park, and Jaegul Choo.
\newblock Stableviton: Learning semantic correspondence with latent diffusion model for virtual try-on.
\newblock In \emph{Proceedings of the IEEE/CVF Conference on Computer Vision and Pattern Recognition}, pages 8176--8185, 2024.

\bibitem[Lab and etc.(2024)]{pku_yuan_lab_and_tuzhan_ai_etc_2024_10948109}
PKU-Yuan Lab and Tuzhan~AI etc.
\newblock Open-sora-plan, 2024.

\bibitem[Lee et~al.(2022)Lee, Gu, Park, Choi, and Choo]{lee2022high}
Sangyun Lee, Gyojung Gu, Sunghyun Park, Seunghwan Choi, and Jaegul Choo.
\newblock High-resolution virtual try-on with misalignment and occlusion-handled conditions.
\newblock In \emph{European Conference on Computer Vision}, pages 204--219. Springer, 2022.

\bibitem[Ma et~al.(2024)Ma, Wang, Jia, Chen, Liu, Li, Chen, and Qiao]{ma2024latte}
Xin Ma, Yaohui Wang, Gengyun Jia, Xinyuan Chen, Ziwei Liu, Yuan-Fang Li, Cunjian Chen, and Yu Qiao.
\newblock Latte: Latent diffusion transformer for video generation.
\newblock 2024.

\bibitem[Morelli et~al.(2022)Morelli, Fincato, Cornia, Landi, Cesari, and Cucchiara]{morelli2022dress}
Davide Morelli, Matteo Fincato, Marcella Cornia, Federico Landi, Fabio Cesari, and Rita Cucchiara.
\newblock Dress code: high-resolution multi-category virtual try-on.
\newblock In \emph{Proceedings of the IEEE/CVF Conference on Computer Vision and Pattern Recognition}, pages 2231--2235, 2022.

\bibitem[Morelli et~al.(2023)Morelli, Baldrati, Cartella, Cornia, Bertini, and Cucchiara]{morelli2023ladi}
Davide Morelli, Alberto Baldrati, Giuseppe Cartella, Marcella Cornia, Marco Bertini, and Rita Cucchiara.
\newblock {LaDI-VTON: Latent Diffusion Textual-Inversion Enhanced Virtual Try-On}.
\newblock In \emph{Proceedings of the ACM International Conference on Multimedia}, 2023.

\bibitem[Oquab et~al.(2024)Oquab, Darcet, Moutakanni, Vo, Szafraniec, Khalidov, Fernandez, Haziza, Massa, El-Nouby, Assran, Ballas, Galuba, Howes, Huang, Li, Misra, Rabbat, Sharma, Synnaeve, Xu, Jegou, Mairal, Labatut, Joulin, and Bojanowski]{oquab2024dinov2learningrobustvisual}
Maxime Oquab, Timothée Darcet, Théo Moutakanni, Huy Vo, Marc Szafraniec, Vasil Khalidov, Pierre Fernandez, Daniel Haziza, Francisco Massa, Alaaeldin El-Nouby, Mahmoud Assran, Nicolas Ballas, Wojciech Galuba, Russell Howes, Po-Yao Huang, Shang-Wen Li, Ishan Misra, Michael Rabbat, Vasu Sharma, Gabriel Synnaeve, Hu Xu, Hervé Jegou, Julien Mairal, Patrick Labatut, Armand Joulin, and Piotr Bojanowski.
\newblock Dinov2: Learning robust visual features without supervision, 2024.

\bibitem[Pont-Tuset et~al.(2018)Pont-Tuset, Perazzi, Caelles, Arbeláez, Sorkine-Hornung, and Gool]{ponttuset20182017davischallengevideo}
Jordi Pont-Tuset, Federico Perazzi, Sergi Caelles, Pablo Arbeláez, Alex Sorkine-Hornung, and Luc~Van Gool.
\newblock The 2017 davis challenge on video object segmentation, 2018.

\bibitem[Raffel et~al.(2023)Raffel, Shazeer, Roberts, Lee, Narang, Matena, Zhou, Li, and Liu]{raffel2023exploringlimitstransferlearning}
Colin Raffel, Noam Shazeer, Adam Roberts, Katherine Lee, Sharan Narang, Michael Matena, Yanqi Zhou, Wei Li, and Peter~J. Liu.
\newblock Exploring the limits of transfer learning with a unified text-to-text transformer, 2023.

\bibitem[Unterthiner et~al.(2018)Unterthiner, van Steenkiste, Kurach, Marinier, Michalski, and Gelly]{unterthiner2018towards}
Thomas Unterthiner, Sjoerd van Steenkiste, Karol Kurach, Raphael Marinier, Marcin Michalski, and Sylvain Gelly.
\newblock Towards accurate generative models of video: A new metric \& challenges.
\newblock \emph{arXiv preprint arXiv:1812.01717}, 2018.

\bibitem[Wang et~al.(2018{\natexlab{a}})Wang, Zheng, Liang, Chen, Lin, and Yang]{wang2018characteristicpreservingimagebasedvirtualtryon}
Bochao Wang, Huabin Zheng, Xiaodan Liang, Yimin Chen, Liang Lin, and Meng Yang.
\newblock Toward characteristic-preserving image-based virtual try-on network, 2018{\natexlab{a}}.

\bibitem[Wang et~al.(2018{\natexlab{b}})Wang, Zheng, Liang, Chen, Lin, and Yang]{wang2018toward}
Bochao Wang, Huabin Zheng, Xiaodan Liang, Yimin Chen, Liang Lin, and Meng Yang.
\newblock Toward characteristic-preserving image-based virtual try-on network.
\newblock In \emph{Proceedings of the European conference on computer vision (ECCV)}, pages 589--604, 2018{\natexlab{b}}.

\bibitem[Wang et~al.(2004)Wang, Bovik, Sheikh, and Simoncelli]{wang2004image}
Zhou Wang, Alan~C Bovik, Hamid~R Sheikh, and Eero~P Simoncelli.
\newblock Image quality assessment: from error visibility to structural similarity.
\newblock \emph{IEEE transactions on image processing}, 13\penalty0 (4):\penalty0 600--612, 2004.

\bibitem[Xu et~al.(2024{\natexlab{a}})Xu, Gu, Chen, and Chen]{xu2024ootdiffusion}
Yuhao Xu, Tao Gu, Weifeng Chen, and Chengcai Chen.
\newblock Ootdiffusion: Outfitting fusion based latent diffusion for controllable virtual try-on.
\newblock \emph{arXiv preprint arXiv:2403.01779}, 2024{\natexlab{a}}.

\bibitem[Xu et~al.(2024{\natexlab{b}})Xu, Chen, Wang, Xing, Zhai, Sang, Lan, Xiao, and Gao]{xu2024tunnel}
Zhengze Xu, Mengting Chen, Zhao Wang, Linyu Xing, Zhonghua Zhai, Nong Sang, Jinsong Lan, Shuai Xiao, and Changxin Gao.
\newblock Tunnel try-on: Excavating spatial-temporal tunnels for high-quality virtual try-on in videos.
\newblock \emph{arXiv preprint}, 2024{\natexlab{b}}.

\bibitem[Yang et~al.(2019)Yang, Fan, and Xu]{yang2019videoinstancesegmentation}
Linjie Yang, Yuchen Fan, and Ning Xu.
\newblock Video instance segmentation, 2019.

\bibitem[Ye et~al.(2023)Ye, Zhang, Liu, Han, and Yang]{ye2023ip-adapter}
Hu Ye, Jun Zhang, Sibo Liu, Xiao Han, and Wei Yang.
\newblock Ip-adapter: Text compatible image prompt adapter for text-to-image diffusion models.
\newblock In \emph{arXiv preprint arxiv:2308.06721}, 2023.

\bibitem[Zhang et~al.(2018)Zhang, Isola, Efros, Shechtman, and Wang]{zhang2018unreasonableeffectivenessdeepfeatures}
Richard Zhang, Phillip Isola, Alexei~A. Efros, Eli Shechtman, and Oliver Wang.
\newblock The unreasonable effectiveness of deep features as a perceptual metric, 2018.

\bibitem[Zhang et~al.(2024)Zhang, Wu, Wang, Luo, Zhang, Zhao, Vajda, Metaxas, and Yu]{zhang2024avid}
Zhixing Zhang, Bichen Wu, Xiaoyan Wang, Yaqiao Luo, Luxin Zhang, Yinan Zhao, Peter Vajda, Dimitris Metaxas, and Licheng Yu.
\newblock Avid: Any-length video inpainting with diffusion model.
\newblock In \emph{Proceedings of the IEEE/CVF Conference on Computer Vision and Pattern Recognition}, pages 7162--7172, 2024.

\bibitem[Zhong et~al.(2021)Zhong, Wu, Tan, Lin, and Wu]{zhong2021mv}
Xiaojing Zhong, Zhonghua Wu, Taizhe Tan, Guosheng Lin, and Qingyao Wu.
\newblock Mv-ton: Memory-based video virtual try-on network.
\newblock In \emph{Proceedings of the 29th ACM International Conference on Multimedia}, pages 908--916, 2021.

\bibitem[Zhu et~al.(2023{\natexlab{a}})Zhu, Yang, Zhu, Reda, Chan, Saharia, Norouzi, and Kemelmacher-Shlizerman]{zhu2023tryondiffusion}
Luyang Zhu, Dawei Yang, Tyler Zhu, Fitsum Reda, William Chan, Chitwan Saharia, Mohammad Norouzi, and Ira Kemelmacher-Shlizerman.
\newblock Tryondiffusion: A tale of two unets.
\newblock In \emph{Proceedings of the IEEE/CVF Conference on Computer Vision and Pattern Recognition}, pages 4606--4615, 2023{\natexlab{a}}.

\bibitem[Zhu et~al.(2023{\natexlab{b}})Zhu, Feng, Chen, Bao, Wang, Chen, Yuan, and Hua]{zhu2023designingbetterasymmetricvqgan}
Zixin Zhu, Xuelu Feng, Dongdong Chen, Jianmin Bao, Le Wang, Yinpeng Chen, Lu Yuan, and Gang Hua.
\newblock Designing a better asymmetric vqgan for stablediffusion, 2023{\natexlab{b}}.

\end{thebibliography}
